\documentclass{article}
\usepackage[utf8]{inputenc}
\usepackage{textgreek}

 \PassOptionsToPackage{numbers, compress}{natbib}

\usepackage[final]{neurips_2025}

\usepackage{hyperref}       
\usepackage{url}            
\usepackage{booktabs}       
\usepackage{amsfonts}       
\usepackage{nicefrac}       
\usepackage{microtype}      
\usepackage[table]{xcolor}
\usepackage{xcolor}         

\usepackage{multirow} 
\usepackage{subfigure}
\usepackage{makecell}
\usepackage{amsmath}
\usepackage{bbding}
\usepackage{subcaption}
\usepackage{adjustbox}
\usepackage{subcaption}
\usepackage{caption}
\usepackage{array}

\title{PoseCrafter: Extreme Pose Estimation with 

Hybrid Video Synthesis}

\author{%
  Qing Mao$^{1,2}$\thanks{Work fully done while first author is a visiting PhD student at the National University of Singapore.}\!\!\quad Tianxin Huang$^3$\!\!\quad Yu Zhu$^1$\!\!\quad Jinqiu Sun$^4$\!\!\quad Yanning Zhang$^1$\thanks{Corresponding author.}\!\!\quad Gim Hee Lee$^2$\footnotemark[2]\\
  $^1$School of Computer Science, Northwestern Polytechnical University\\
  $^2$School of Computing, National University of Singapore\\
  $^3$School of Computing and Data Science, The University of Hong Kong\\
  $^4$School of Astronautics, Northwestern Polytechnical University\\
  \texttt{maoqing@mail.nwpu.edu.cn \quad ynzhang@nwpu.edu.cn \quad gimhee.lee@nus.edu.sg} \\
    {\tt \href{https://github.com/maoqingsunny/PoseCrafter}{\textbf{https://github.com/maoqingsunny/PoseCrafter}}}
}

\begin{document}

\maketitle

\begin{abstract}
Pairwise camera pose estimation from sparsely overlapping image pairs remains a critical and unsolved challenge in 3D vision. 
Most existing methods struggle with image pairs that have small or no overlap. Recent approaches attempt to address this by synthesizing intermediate frames using video interpolation and selecting key frames via a self-consistency score. However, the generated frames are often blurry due to small overlap inputs, and the selection strategies are slow and not explicitly aligned with pose estimation.
To solve these cases, we propose \textbf{Hybrid Video Generation (HVG)} to synthesize clearer intermediate frames by coupling a video interpolation model with a pose-conditioned novel view synthesis model, where we also propose a \textbf{Feature Matching Selector (FMS)} based on feature correspondence to select intermediate frames appropriate for pose estimation from the synthesized results.
Extensive experiments on Cambridge Landmarks, ScanNet, DL3DV-10K, and NAVI demonstrate that, compared to existing SOTA methods, PoseCrafter can obviously enhance the pose estimation performances, especially on examples with small or no overlap.
\end{abstract}

\section{Introduction}
Accurately estimating the relative pose between two images is a fundamental task in 3D vision, critical for enabling autonomous navigation, robotics, and augmented reality. While modern feature-based~\citep{sift, surf} or learning-based pipelines~\citep{ DUSt3R,mast3r}perform well when input views share sufficient overlap, their performance deteriorates sharply in the small or no overlap case, owing to the fundamental scarcity of reliable feature correspondences.

To address this issue, InterPose~\citep{InterPose} introduced a two-stage strategy: intermediate views are generated using a video interpolation model, and enriched frames selected with a frame selection strategy are provided to the pose estimation model. Although InterPose demonstrated obvious improvements in extreme cases with small overlaps, it faces two core limitations. 
First, the quality of generated intermediate frames remains a bottleneck. 
Free and open-source models like DynamiCrafter~\citep{dynamicrafter} can generate plausible interpolations near the input image pair but often fail to maintain geometric consistency in the central frames, likely due to the limited overlap between input views.
Commercial models like Runway~\citep{Runway} and Luma Dream Machine~\citep{LumaAI} offer sharper results but still suffer from blur and drift in challenging scenes, in addition to incurring high inference costs. 
Second, InterPose relies on a self-consistency score mechanism to select frames, which filters out informative frames across multiple generated video clips. This purely statistical selection strategy is slow and not explicitly aligned with the objectives of pose estimation.

To address these challenges, we propose \textbf{PoseCrafter}—a training-free framework for estimating camera poses from input image pairs with small or no overlap. PoseCrafter comprises two key components: Hybrid Video Generation (HVG) and Feature Matching Selector (FMS).
In HVG, we first use the pre-trained video interpolation model DynamiCrafter~\citep{dynamicrafter} to synthesize a coarse video and identify a few reliable “relay” frames. These selected frames are then used to estimate the initial pose and generate high-fidelity intermediate views through a pose-conditioned novel view synthesis model, ViewCrafter~\citep{viewcrafter}.
In this work, we find that selecting only the one frame immediately after the start frame and the one frame just before the end frame yields the most effective relay frames. Including additional frames tends to degrade performance due to increased blurriness in the central synthesized frames. More analysis can be found in  Sec.~\ref{sec:relay}.

In FMS, instead of using statistical scores for frame selection, we propose a simple but effective solution, evaluating synthesized frames based on their feature correspondences with the input image pair to determine their suitability for pose estimation.
In detail, we extract local descriptors from each candidate frame and match them to the input image pair, compute RANSAC~\citep{ransac} inlier counts, and select the top \(k\) frames with the highest total inliers.

Extensive experiments on common benchmarks, including Cambridge Landmarks~\citep{cambridgelandmarks}, ScanNet~\citep{scannet}, DL3DV-10K~\citep{dl3dv}, and NAVI~\citep{navi} show that PoseCrafter can obviously improve the accuracy of pose estimation on extreme pose image pairs with small or no overlaps, without any requirements for additional training or ground-truth supervision.
In summary, our contributions are as follows:
\begin{itemize}
  \item We propose \textbf{Hybrid Video Generation(HVG)} to synthesize high-fidelity intermediate frames, by effectively coupling a video interpolation model and a pose-conditioned novel view synthesis model.
  \item We develop a \textbf{Feature Matching Selector(FMS)} to deterministically select the most informative frames suitable for pose estimation, eliminating the need for expensive statistical self‐consistency scoring in existing works.
  \item We demonstrate that, by integrating HVG and FMS, our proposed \textbf{PoseCrafter} can achieve SOTA performances across four challenging benchmarks for extreme pose estimation.
\end{itemize}

\section{Related Work}
\paragraph{Extreme Pose Estimation}
Traditional relative pose pipelines combine local feature matching \citep{sift, surf}, RANSAC-based~\citep{ransac} five-point or eight-point solvers, and bundle adjustment~\citep {colmap}. These methods perform well when images share substantial overlap but fail under extreme cases. While deep networks have strengthened individual components~\citep{loftr,patch2pix,relpose,sparsepose}, they still rely on the large overlap. To address cases with small or no overlap, several approaches~\citep{yangextreme,caiextreme,bezalel2024extreme, DUSt3R} have been proposed. Yang et al.~\citep{yangextreme} alternate RGB-D scene completion with pose regression, iteratively fusing geometry from both scans; Cai et al.~\citep{caiextreme} construct dense correlation volumes over synthetic video clips to infer large rotations; and  DUSt3R~\citep{ DUSt3R} regresses pixel-aligned point maps to jointly recover depth, reconstruction, and pose without camera calibration.  Temporal modeling has also been explored. Tang et al.~\citep{rnnpose} combine attention and recurrent networks for pose prediction. However, each of these methods requires supervised data and training resources. More recently, video‐diffusion techniques have been adapted for pose recovery. JOG3R~\citep{jog3r} fine‐tunes intermediate features of a pre-trained video generation model for SfM‐style estimation, InterPose~\citep{InterPose} reuses standard diffusion interpolation and a self‐consistency scoring scheme to select the most reliable frames, both without retraining the video model. These studies highlight the power of synthesized intermediate views and motivate our PoseCrafter framework.
\paragraph{Generative Video Models }
Early video synthesis methods relied on generative adversarial networks (GANs)~\citep{gan} and autoregressive models. GAN‐based approaches such as TGAN~\citep{tgan}, MoCoGAN~\citep{mocogan}, and StyleGAN‐V~\citep{styleganv} extend image GANs to model temporal coherence, but often suffer from frame inconsistencies and require complex architectures. Autoregressive techniques~\citep{vpn,latentvideo}generate videos frame by frame, but they incur high computational costs and error accumulation. Diffusion models have recently set a new standard for both image and video synthesis. Video‐diffusion frameworks such as VDM~\citep{vdm}, Imagen‐Video~\citep{cascaded}, and Make‐A‐Video~\citep{make} generate high‐fidelity, temporally coherent sequences via iterative denoising. ViewCrafter~\citep{viewcrafter} proposed by combining video diffusion priors with coarse point-cloud cues to synthesize high-fidelity, pose-controlled novel views from one or a few images. There are also some video generation models that can be used for interpolation that use diffusion to generate smooth, geometry-aware intermediate frames. For example, DynamiCrafter~\citep{dynamicrafter} is based on motion-consistent editing. These diffusion‐based models outperform prior GAN and autoregressive methods in both visual quality and temporal consistency, making them ideal for downstream tasks such as extreme pose estimation.
\begin{figure}
            \centering
            \includegraphics[width=1\linewidth]{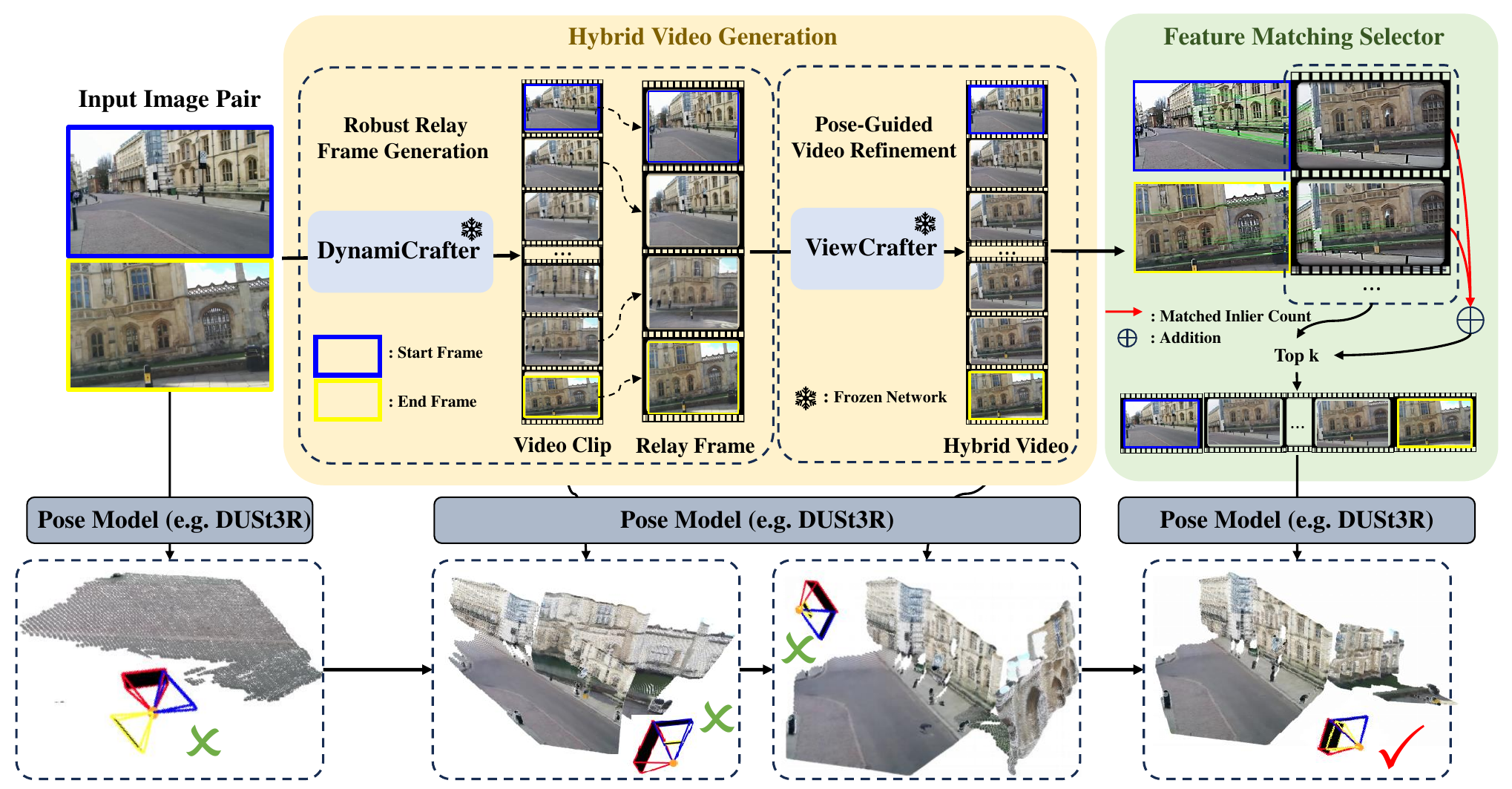}
            \caption{Overview of the PoseCrafter pipeline. Starting from an input image pair (left), the \textbf{Hybrid Video Generation} module uses DynamiCrafter to interpolate a short video clip, extracts the most reliable “relay” frames, and refines the sequence using the pose-conditioned ViewCrafter to reduce mid-sequence blur and drift. The \textbf{Feature Matching Selector} then evaluates each synthesized frame based on its feature correspondence with the input pair, selecting the most informative views for pose estimation. These selected frames are then fed into a pose estimation model (e.g.  DUSt3R) to estimate the final pose.
            }
            \label{fig:main}
\end{figure}
\section{Our Method}
As illustrated in Figure~\ref{fig:main}, given a pair of images with small or no overlap, \textbf{PoseCrafter} estimates their relative camera pose through two complementary stages. In \textbf{Hybrid Video Generation} (Section~\ref{Hybrid video generation}), we sequentially combine a video interpolation model, DynamiCrafter, with a pose-conditioned novel view synthesis model, ViewCrafter, to generate refined video frames that significantly reduce mid-sequence blurring and correct structural drift. In the subsequent \textbf{Feature Matching Selector} (Section~\ref{Feature‐Matching‐Based Frame Selection}), we evaluate the synthesized frames based on their feature correspondences with the input image pair and select the top \(k\) most informative frames for final pose estimation.

\subsection{Preliminaries}
\vspace{-0.1in}
\paragraph{DynamiCrafter}
DynamiCrafter is a publicly available diffusion model for video synthesis that employs a progressive latent denoising architecture to animate two static images into dynamic sequences~\citep{dynamicrafter}. 
Given an input image pair (\(I_0, I_T \in \mathbb{R}^{H\times W\times 3}\)) and an optional text prompt \(\mathcal{P}\), it generates a dense sequence of intermediate frames using the following formula:
\[
\{I_t\}_{t=0}^T = \mathcal{G}_{DC}(I_0, I_T, \mathcal{P}),
\]
where \( \mathcal{G}_{DC} \) denotes the pre-trained DynamiCrafter model. The variable \( t \in \mathbb{Z}^+ \) denotes the frame index, where \( I_0 \) and \( I_T \) correspond to the input image pair. 

\vspace{-0.1in}
\paragraph{ViewCrafter}
ViewCrafter is a pose-conditioned diffusion model for novel view synthesis of static 3D scenes, which can generate high-fidelity novel views along an interpolated camera trajectory through input images following the formula:  
\[
\{I_t\}_{t=0}^T = \mathcal{G}_{VC}(I_0, I_T, \{C_t\}_{t=0}^T),
\]
where \( \mathcal{G}_{VC} \) denotes the pre-trained ViewCrafter generator. $\{C_t\}_{t=0}^T$ denotes camera poses alongside the trajectory interpolated between the input image pair.

\subsection{Hybrid Video Generation}
\label{Hybrid video generation}
\begin{figure}
            \centering
            \includegraphics[width=1\linewidth]{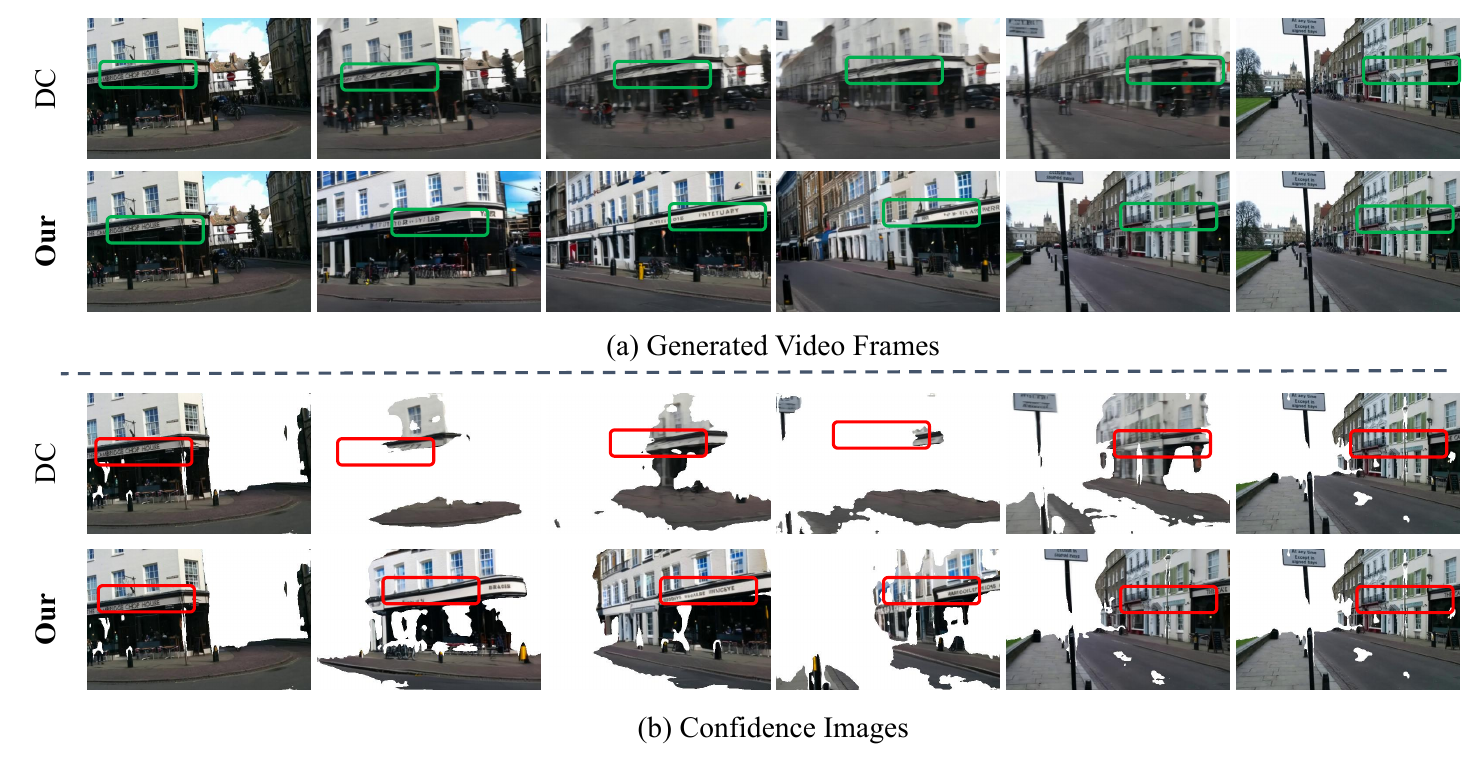}
            \caption{Examples of mid-frame blurriness and confidence degradation. (a) presents frames generated by DynamiCrafter (DC) and our proposed method. (b) highlights high-confidence regions identified by DUST3R during pose estimation. \textcolor{green}{Green} boxes mark areas affected by blurriness in the DC-generated frames, while \textcolor{red}{red} boxes indicate how these regions performed in DUST3R’s confidence maps. We observe that the corresponding regions in the DynamiCrafter results exhibit notably low confidence, whereas our method generates frames with higher confidence for DUST3R.}
            \label{fig:confidence}
\end{figure}

Although InterPose~\citep{InterPose} improves pose estimation performance by directly interpolating frames between input image pairs (e.g., using DynamiCrafter~\citep{dynamicrafter}), the generated sequences often exhibit noticeable blurriness and inconsistencies, particularly in the middle frames. As shown in Figure~\ref{fig:confidence}, these blurry regions result in significantly lower confidence during pose estimation with  DUSt3R~\citep{ DUSt3R}, ultimately affecting overall performance.
In contrast, ViewCrafter~\citep{viewcrafter} is capable of synthesizing clear, high-fidelity frames, but it requires input images that allow for the estimation of a plausible camera trajectory.
To address this issue, we propose Hybrid Video Generation (HVG), which couples DynamiCrafter and ViewCrafter. Specifically, we first use DynamiCrafter to generate initial interpolated frames and then select a few reliable “relay” frames. These relay frames help ensure a plausible camera trajectory and are subsequently used by ViewCrafter to synthesize high-quality intermediate frames.
As shown in Figure~\ref{fig:confidence}-Our, intermediate frames generated by our HVG exhibit higher confidence scores in  DUSt3R~\citep{ DUSt3R}, further confirming that HVG contributes positively to pose estimation.
The whole process comprises two phases: Robust Relay-Frame Generation, and Pose-Guided Video Refinement.
\vspace{-0.1in}
\paragraph{Robust Relay-Frame Generation}
\label{sec:relay}
In this stage, we use the pre-trained DynamiCrafter model to interpolate a short video clip from the input image pair (the start frame \(I_0\) and the end frame \(I_T\)). 
Although DynamiCrafter generates plausible interpolations near the input pair, it often fails to maintain geometric consistency in the central frames, weakening feature correspondences and degrading downstream pose estimation. 
To isolate the most reliable “relay” frames, we conducted a study (Table~\ref{tab:Rfg}) that evaluated various subsets of interpolated frames. 
We found that retaining only the start/end frames and their immediate neighbors \(\{I_0, I_1, I_{T-1}, I_T\}\) minimizes rotation error and maximizes stability. 
Consequently, PoseCrafter forwards only these four frames as “relay” frames to the subsequent high-fidelity novel view synthesis stage.
\vspace{-0.1in}
\paragraph{Pose‐Guided Video Refinement}
Given the four reliable relay frames \(\{I_0,I_1,I_{T-1},I_{T}\}\), we first recover their camera poses \(\{C_0,C_1,C_{T-1},C_{T}\}\) using the pretrained  DUSt3R network. We then interpolate a smooth, dense camera trajectory \(\{C_t\}_{t=0}^{T}\) by applying spherical linear interpolation to rotations on \(\mathrm{SO}(3)\) and linear interpolation to translations in \(\mathbb{R}^3\). Conditioned on this trajectory and the reliable relay frames, ViewCrafter synthesizes high‐fidelity intermediate views.
This pose‐conditioned refinement effectively removes mid‐sequence blur and geometric drift, producing crisp, structurally accurate frames that markedly improve feature correspondences for final pose estimation.

\begin{table}[!ht]
    \caption{Relay-frame sampling analysis using mean rotation error (MRE\(\downarrow\)). The setting \#Frames=2 corresponds to \(\{I_0,I_T\}\), \#Frames=4 corresponds to \(\{I_0,I_1,I_{T-1},I_T\}\), \#Frames=6 corresponds to \(\{I_0,I_1,I_2,I_{T-2},I_{T-1},I_T\}\), and \#Frames=8 corresponds to \(\{I_0,I_1,I_2,I_3,I_{T-3},I_{T-2},I_{T-1},I_T\}\). The case \#Frames=16 uses all frames. Results indicate that \#Frames=4 consistently achieves the lowest MRE and the highest stability across datasets. }
  \label{tab:Rfg}
\centering
\renewcommand{\arraystretch}{1.2}
\begin{tabular}{lccccc}
\toprule
\multirow{2}{*}{\textbf{Dataset}} & \multicolumn{5}{c}{\textbf{\#Frames ($n$)}} \\
\cmidrule(lr){2-6}
                 & 2     & 4     & 6     & 8     & 16    \\
\midrule
\textbf{Cambridge Landmarks} & 20.56 & \textbf{14.47} & 16.66 & 16.87 & 17.83 \\
\textbf{ScanNet}             & 19.67 & \textbf{16.23} & 17.03 & 17.16 & 18.56 \\
\textbf{DL3DV-10K}           & 15.22 & \textbf{14.27} & 14.40 & 14.73 & 14.52 \\
\textbf{NAVI}                &  7.78 & \textbf{6.94}  &  7.18 &  9.64 & 10.92 \\
\bottomrule
\end{tabular}
\vspace{0.5em}
\end{table}

\subsection{Feature Matching Selector} 
\label{Feature‐Matching‐Based Frame Selection}
Although our proposed Hybrid Video Generation (HVG) produces high-fidelity and visually clear intermediate frames, not all of them are equally beneficial for pose estimation, as some generated content may be inconsistent with the input image pair.
InterPose~\citep{InterPose} addresses this problem by computing a statistical self-consistency score across multiple separate estimations. However, this approach is both slow and unstable due to the inherent uncertainty introduced by repeated sampling. Moreover, as a purely statistical measure, it does not explicitly ensure that the selected frames are well-suited for pose estimation.

To overcome these drawbacks, we employ a simple but effective deterministic selection strategy based on feature matching, named as Feature Matching Selector (FMS). For each synthesized frame \(I_t\), we extract local descriptors and match them to both the start frame \(I_0\) and end frame \(I_T\), then compute RANSAC inlier counts \(N_0(t)\) and \(N_T(t)\). Each frame is scored by  
\[
S(t) = N_0(t) + N_T(t),
\]  
and we select the top \(k\) frames whose scores exceed a preset threshold. 
This inlier-driven criterion avoids the need for repeated estimations and explicitly accounts for geometric consistency with the input image pair, enabling the selection of more informative frames for subsequent pose estimation.

\section{Experiments}
\subsection{Experiment Setup}
\textbf{Data Preparation}
Following InterPose~\citep{InterPose}, we evaluate our method on four common benchmarks:
Cambridge Landmarks~\citep{cambridgelandmarks}, ScanNet~\citep{scannet}, Navi~\citep{navi} and DL3DV-10K\citep{dl3dv}.  For Cambridge Landmarks and ScanNet, we select test pairs by sampling images whose relative yaw difference falls into two ranges ( \([50^\circ\text{--}65^\circ]\) and \([65^\circ\text{--}90^\circ]\)) to evaluate performance under small and no overlap cases. For the object-centric Navi and DL3DV-10K datasets, due to the large object overlap, we adopt a single yaw range of \([50^\circ\text{--}90^\circ]\) following InterPose's setting. 
All images are resized, and center-cropped to \(512 \times 320 \) before processing, to adapt to the setting of DynamiCrafter~\citep{dynamicrafter}.

\textbf{Evaluation Metrics}
Following prior work~\citep{InterPose}, we conduct evaluations using the following metrics:
\vspace{-0.15in}
\paragraph{Mean Rotation Error (MRE)}
The average geodesic distance between the predicted and the ground‐truth rotation, reported in angular.
\vspace{-0.15in}
\paragraph{Mean Translation Error (MTE)}  
The average geodesic distance between the predicted and the ground‐truth translation, reported in angular.
\vspace{-0.15in}
\paragraph{Rotation and Translation Accuracy (\(R(\theta),T(\theta)\))}
 Given an angular threshold \(\theta\) (e.g., \(5^\circ\), \(15^\circ\), \(30^\circ\)), the rotation accuracy \(R(\theta)\)  is the fraction of estimates with rotation error below \(\theta\); the translation accuracy \(T(\theta)\) is the same definition using translation angular error.
\vspace{-0.15in}
\paragraph{Area Under Curve (AUC\(_{30})\)}
The normalized area under the rotation or translation error curve from \(0^\circ\) to \(30^\circ\) provides a compact summary of performance across the full range of angular thresholds. In this work, we report the smaller of the rotation and translation AUC\(_{30}\) values as the final metric.
\subsection{Implementation Details}
In the hybrid video generation stage, we first generate 16 interpolated frames between each input image pair using DynamiCrafter. From these, we select 4 frames as reliable relay frames (as described in section \ref{sec:relay}), which are then used by ViewCrafter to render 25-frame sequences for subsequent selection with Feature Matching Selector (FMS).
In FMS, we extract 2,000 ORB keypoints per frame and compute RANSAC inlier counts with respect to both the start and end keyframes. The generated frames are then ranked based on their inlier counts, and the top \(k=6\) frames(excluding the input image pair) are selected for the final pose estimation. All experiments were conducted on a single NVIDIA RTX 6000 GPU to ensure a consistent evaluation environment.

\subsection{Comparison with State-of-the-Art}
To validate our method's effectiveness, we compare it against two closely related baselines:  DUSt3R~\citep{ DUSt3R} and InterPose~\citep{InterPose}.
DUSt3r estimates relative pose directly from an input image pair,
while InterPosee enhances pose estimation results by using videos generated by a pre-trained video model and intermediate frames selected by a self-consistency score.
Since InterPose has not been publicly available, we reproduce its pipeline based on the details provided in their paper, and denote it as InterPose\(^{\ddagger}\) for subsequent comparisons.
To further assess the impact of video generation quality, we also introduce two variant methods:
our method without feature matching selector(Ours\(_{\text{w/o FMS}}\)), and InterPose without self-consistency score (InterPose\(_{\text{w/o SCS}}^\ddagger\)). 
In these cases, all of the synthesized frames are used for pose estimation, without specific selection.

\subsubsection{Quantitative Comparison on PoseCrafter}
Tables~\ref{tab:cambridge}, \ref{tab:scannet}, \ref{tab:dl3dv}, and \ref{tab:navi} present comparative performance results of PoseCrafter against other methods across four benchmark datasets. 
On outward‐facing benchmarks (Cambridge Landmarks and ScanNet), PoseCrafter reduces mean rotation error by approximately \(9.85^\circ\) and increases R@30\(^\circ\) by more than \(10\%\) against InerPose\(^{\ddagger}\) for data with yaw changes between \([65^\circ\text{--}90^\circ]\) of Cambridge Landmarks.
These improvements demonstrate our pipeline’s superiority on extreme pose estimation.
Moreover, the comparison between \(\text{InterPose}_{\text{w/o SCS}}^{\ddagger}\) and \(\text{PoseCrafter}_{\text{w/o FMS}}\) further confirms that our proposed hybrid video generation approach produces frames more suitable for pose estimation than the DynamiCrafter~\citep{dynamicrafter} used in InterPose~\citep{InterPose}.

On center‐facing benchmarks (DL3DV‐10K and NAVI) with larger cross-image overlaps, PoseCrafter still yields modest but consistent reductions in rotation and translation error, alongside incremental improvements in \(\text{AUC}_{30}\).
Overall, the statistical results across multiple standard benchmarks confirm that our proposed pipeline based on Hybrid Video Generation and the Feature Matching Selector, provides a stable, training-free solution across diverse scenes and baseline conditions.
\begin{table*}[!ht]
  \centering
  \caption{Pose estimation on Cambridge Landmarks. We report rotation recall (R@\(\,\theta\uparrow\)), translation recall (T@\(\,\theta\uparrow\)), mean rotation error (MRE\(\downarrow\)), and AUC\(_{30}\uparrow\).}
 \resizebox{\linewidth}{!}{\label{tab:cambridge}
  \begin{tabular}{@{} ll  *{5}{c}  *{5}{c} @{}}
    \toprule
    \multirow{2}{*}{\textbf{Method}}  & 
    \multirow{2}{*}{\textbf{Input}}   & 
    \multicolumn{5}{c}{\textbf{Yaw range} [50\(^{\circ}\)-65\(^{\circ}\)]}
                             & \multicolumn{5}{c}{ \textbf{Yaw range} [65\(^{\circ}\)-90\(^{\circ}\)]} \\
    \cmidrule(lr){3-7} \cmidrule(l){8-12}
                             & 
                             & \textbf{MRE\(\downarrow\)}  
                             & \textbf{R@5\(^\circ\)}
                             & \textbf{R@15\(^\circ\)} 
                             & \textbf{R@30\(^\circ\)}
                             & \textbf{AUC\(_{30}\uparrow\)}
                             & \textbf{ MRE\(\downarrow\)}  
                             & \textbf{R@5\(^\circ\)} 
                             & \textbf{R@15\(^\circ\)} 
                             & \textbf{R@30 }
                             & \textbf{AUC\(_{30}\uparrow\)} \\
    \midrule
     DUSt3R                    & Pair       & 18.14 & 40.34 & 71.25 & 82.99 & 61.98 & 51.24 & 21.67 & 44.67 & 51.67 & 37.93 \\
    InterPose\(_{\text{w/o SCS}}^\ddagger\) & DynamiCrafter & 16.11 & 42.70 & 75.70 & 87.35 & 65.72 & 42.51 & 30.67 & 42.51 & 61.33 & 47.18 \\
    InterPose\(^{\ddagger}\)                  & DynamiCrafter & 13.61 & 51.81 & 81.50 & 83.30 & 70.47 & 38.87 & 36.33 & 65.67 & 68.33 & 55.24 \\
     Ours\(_{\text{w/o FMS}}\)          & Hybrid video  & 13.24 & 54.51 & 89.24 & 92.71 & 76.13 & 34.87 & 31.33 & 68.33 & 77.67 & 56.29 \\
   Ours         & Hybrid video  & \textbf{11.40} & \textbf{55.21} & \textbf{89.93} & \textbf{93.75} & \textbf{77.41} & \textbf{29.02} & \textbf{36.67} & \textbf{71.67} & \textbf{78.33} & \textbf{60.46} \\
    \bottomrule
  \end{tabular}}
\end{table*}

\begin{table*}[!ht]
\centering
\caption{Pose estimation on ScanNet. We report rotation recall (R@\(\,\theta\uparrow\)), translation recall (T@\(\,\theta\uparrow\)), mean rotation error (MRE\(\downarrow\)), mean translation error (MTE\(\downarrow\)), and AUC\(_{30}\uparrow\).}
\label{tab:scannet}
\resizebox{\linewidth}{!}{
\begin{tabular}{l l c c c c c c c c c c c}
\toprule
\textbf{Yaw range} & \textbf{Method} & \textbf{Input} & \textbf{R@5\(^\circ\)} & \textbf{R@15\(^\circ\)} & \textbf{R@30\(^\circ\)} & \textbf{T@5°} & \textbf{T@15°} & \textbf{T@30°} & \textbf{MRE↓} & \textbf{MTE↓} & \textbf{AUC30↑} \\
\midrule
\multirow{5}{*}{50\(^{\circ}\)-65\(^{\circ}\)} &  DUSt3R & Pair  & 43.97 & 74.14 & 79.31 & 25.34 & 52.07 & \textbf{78.45} & 19.41 & 25.23 & 47.37 \\
& InterPose\(_{\text{w/o SCS}}^\ddagger\) & DynamicCrafter  & 46.55 & 77.59 & 85.34 & 16.38 & 48.10 & 64.74 & 17.51 & 35.25 & 42.69 \\
& InterPose\(^{\ddagger}\) & DynamicCrafter & 50.86 & 81.03 & 87.07 & 27.58 & 61.21 & 69.46 & 15.15 & 23.89 & 53.33 \\
&  Ours\(_{\text{w/o FMS}}\)\ & Hybrid video & 51.72 & 87.07 & 93.10 & 23.28 & 50.62 & 67.07 & 12.38 & 29.02 & 45.53 \\
& Ours & Hybrid video  & \textbf{53.45} & \textbf{88.79} & \textbf{94.83 }& \textbf{33.62} &\textbf{ 65.52} & 77.69 & \textbf{10.77} & \textbf{22.14} & \textbf{57.03} \\
\midrule
\multirow{5}{*}{65\(^{\circ}\)-90\(^{\circ}\)} &  DUSt3R & Pair  & 42.05 & 67.05 & 70.45 & 26.59 & 46.59 & 53.86 & 30.82 & 29.99 & 36.50 \\
& InterPose\(_{\text{w/o SCS}}^\ddagger\) & DynamicCrafter & 38.64 & 62.50 & 65.90 & 20.45 & 39.77 & 47.72 & 35.18 &58.89 & 33.40 \\
& InterPose & DynamicCrafter & 45.45  & 67.05  & 71.59  &31.81  & 53.41  & 64.77   & 28.22  & 29.52 & 45.98  \\
& Ours\(_{\text{w/o FMS}}\)\ & Hybrid video & 46.59  & 76.14  & 82.95  &23.85  & 48.86  & 57.95  & 22.61  & 35.98& 41.72  \\
& Ours & Hybrid video & \textbf{50.00}     & \textbf{77.72}  & \textbf{84.09}  &\textbf{37.50}   & \textbf{63.64}  & \textbf{73.86}  & \textbf{17.02}  & \textbf{29.28}& \textbf{56.44}    \\
\bottomrule
\end{tabular}
}
\end{table*}

\begin{table}[!ht]
  \caption{Pose estimation on DL3DV-10K with [50\(^{\circ}\)-90\(^{\circ}\)] yaw range. }
  \label{tab:dl3dv}
  \centering
  \resizebox{\linewidth}{!}{
    \begin{tabular}{l l c c c c c c c c c c}
      \toprule
      \textbf{Method} & \textbf{Input} & \textbf{R@5\(^\circ\)} & \textbf{R@15\(^\circ\)} & \textbf{R@30\(^\circ\)} & \textbf{T@5\(^\circ\)} & \textbf{T@15\(^\circ\)} & \textbf{T@30\(^\circ\)} & \textbf{MRE\(\downarrow\)} & \textbf{MTE\(\downarrow\)} & \textbf{AUC\(_{30}\uparrow\)} \\
      \midrule
       DUSt3R                         & Pair            & 34.33 & 63.00 & 94.66 & 27.00 & 75.00 & 92.67 & 13.36 & 10.88 & 55.58 \\
      InterPose\(_{\text{w/o SCS}}^{\ddagger}\) & DynamicCrafter  & 36.33 & 64.33 & 95.00 & 26.00 & 76.33 & 92.67 & 13.32 & 11.27 & 55.68 \\
      InterPose\(^{\ddagger}\)         & DynamicCrafter  & 36.11 & 64.33 & 97.66 & 27.66 & 79.67 & 95.33 & 13.17 & 10.76 & 56.05 \\
      Ours\(_{\text{w/o FMS}}\)        & Hybrid video    & 38.33 & 68.33 & 98.33 & 30.66 & 79.61 & 96.67 & 12.89 & 10.71 & 57.16 \\
      Ours                  & Hybrid video & \textbf{38.10} & \textbf{70.00} & \textbf{100.00} & \textbf{31.33} & \textbf{81.33} & \textbf{98.33} & \textbf{12.73} & \textbf{10.28} & \textbf{57.48} \\
      \bottomrule
    \end{tabular}
  }
\end{table}

\begin{table}[!ht]
  \caption{Pose estimation on NAVI for the [50\(^{\circ}\)-90\(^{\circ}\)] yaw range.}
  \label{tab:navi}
  \centering
  \resizebox{\linewidth}{!}{
    \begin{tabular}{l l c c c c c c c c c c}
      \toprule
      \textbf{Method} & \textbf{Input} & \textbf{R@5\(^\circ\)} & \textbf{R@15\(^\circ\)} & \textbf{R@30\(^\circ\)} & \textbf{T@5\(^\circ\)} & \textbf{T@15\(^\circ\)} & \textbf{T@30\(^\circ\)} & \textbf{MRE\(\downarrow\)} & \textbf{MTE\(\downarrow\)} & \textbf{AUC\(_{30}\uparrow\)} \\
      \midrule
       DUSt3R                         & Pair            & 64.69 & 95.72 & 98.05 & 62.37 & 97.28 & 98.22 & 7.30  & 7.82  & 82.37 \\
      InterPose\(_{\text{w/o SCS}}^{\ddagger}\) & DynamicCrafter  & 45.13 & 92.61 & 96.11 & 57.86 & 91.44 & 96.50 & 11.14 & 8.81  & 78.63 \\
      InterPose\(^{\ddagger}\)         & DynamicCrafter  & 66.53 & 97.28 & \textbf{98.83} & 67.70 & 96.89 & 98.84 & 6.61  & 6.26  & 82.80 \\
      Ours\(_{\text{w/o FMS}}\)        & Hybrid video    & 59.53 & 97.28 & \textbf{98.83} & 72.26 & 95.72 & 98.83 & 6.93  & 6.87  & 81.91 \\
      Ours                  & Hybrid video & \textbf{70.82} & \textbf{97.67} & \textbf{98.83} & \textbf{75.10} & \textbf{98.44} & \textbf{99.22} & \textbf{5.97} & \textbf{5.46} & \textbf{83.98} \\
      \bottomrule
    \end{tabular}
  }
\end{table}

\begin{table}[!ht]
  \caption{Runtime and Memory Cost. }
  \label{tab:cost}
  \centering
  \resizebox{\linewidth}{!}{
    \begin{tabular}{lcclcc}
\toprule
\multirow{2}{*}{\textbf{Method}} & \multicolumn{2}{c}{\textbf{Runtime}} &  & \multicolumn{2}{c}{\textbf{Memory Cost}} \\ \cline{2-3} \cline{5-6} 
                                 & Video Generation      & Pose Estimation      &  & Video Generation       & Pose Estimation       \\ 
                                 \midrule
InterPose\(^{\ddagger}\)               & 3.2min                & 20.29min       &  & 14.6GB                & 3.1GB\\
Ours                    & 3.8min                & 0.18min        &  & 22.8GB                & 3.6GB          
          \\ \bottomrule
\end{tabular}
  }
\end{table}

\begin{table}[!ht]
  \caption{Ablation study on Hybrid Video Generation. SCS and FMS denote the frame selection strategies from InterPose~\citep{InterPose} and ours, respectively. 
  }
  \label{tab:ab1}
  \centering
  \renewcommand{\arraystretch}{0.95}
  \setlength{\tabcolsep}{3pt}
  \scalebox{0.87}{
    \begin{tabular}{l l c c c c c}
      \toprule
      \textbf{Method} & \textbf{Input} & \textbf{MRE\(\downarrow\)} & \textbf{R@5\(^\circ\)} & \textbf{R@15\(^\circ\)} & \textbf{R@30\(^\circ\)} & \textbf{AUC\(_{30}\uparrow\)} \\
      \midrule
       DUSt3R                             & Pair             & 18.14 & 40.34 & 71.25 & 82.99 & 61.98 \\
      InterPose\(_{\text{w/o SCS}}^{\ddagger}\) & DynamicCrafter   & 16.11 & 42.70 & 75.70 & 87.50 & 65.72 \\
      InterPose\(^{\ddagger}\)            & DynamicCrafter   & 13.60 & 51.81 & 81.50 & 83.30 & 70.47 \\
      InterPose\(_{\text{w/ FMS}}^{\ddagger}\)  & DynamicCrafter   & 13.02 & 52.08 & 85.76 & 90.63 & 73.93 \\
      \midrule
      ViewCrafter\(_{\text{w/o FMS}}\)     & ViewCrafter      & 13.80 & 52.78 & 82.29 & 88.54 & 71.12 \\
      ViewCrafter\(_{\text{w/ FMS}}\)      & ViewCrafter      & 12.45 & 53.82 & 84.03 & 90.28 & 72.82 \\
      \midrule
      Ours\(_{\text{w/o FMS}}\)           & Hybrid video     & 13.24 & 54.51 & 89.24 & 92.71 & 76.13 \\
      Ours\(_{\text{w/ SCS}}\)            & Hybrid video     & 12.11 & 54.86 & 88.54 & 91.32 & 76.11 \\
      Ours                     & Hybrid video & \textbf{11.40} & \textbf{55.21} & \textbf{89.93} & \textbf{93.75} & \textbf{77.41} \\
      \bottomrule
    \end{tabular}
  }
\end{table}

\begin{table}[!ht]
  \caption{Ablation study on feature matching methods used in FMS.}
  \label{tab:ab2}
  \centering
  \renewcommand{\arraystretch}{0.95}
  \setlength{\tabcolsep}{4pt}
  \scalebox{0.92}{
    \begin{tabular}{c c c c c c}
      \toprule
      \textbf{Feature Matching Method} & \textbf{MRE\(\downarrow\)} & \textbf{R@5\(^\circ\)} & \textbf{R@15\(^\circ\)} & \textbf{R@30\(^\circ\)} & \textbf{AUC\(_{30}\uparrow\)} \\
      \midrule
      ORB        & \textbf{11.40} & \textbf{55.21} & \textbf{89.93} & \textbf{93.75} & \textbf{77.41} \\
      RoMa   & 13.89 & 53.13 & 88.89 & 91.67 & 74.76\\ 
      LoFTR  & 12.41                   & 54.51                  & 89.24                   & 93.40                   & 76.30                     \\ 
      SIFT                & 13.89          & 53.13          & 88.89          & 92.67          & 74.77 \\
      SuperPoint          & 13.36          & 51.74          & \textbf{89.93}          & 92.71          & 75.48 \\
      \bottomrule
    \end{tabular}
  }
\end{table}

\begin{table}[!ht]
  \caption{Ablation study on different video interpolation models used in Hybrid Video Generation. }
  \label{tab:ab3}
  \centering
  \renewcommand{\arraystretch}{0.95}
  \setlength{\tabcolsep}{4pt}
  \scalebox{0.92}{
    \begin{tabular}{c c c c c c c}
      \toprule
     \textbf{Method} &\textbf{Video Interpolation Model} & \textbf{MRE↓}   & \textbf{R@5\(^\circ\)}    & \textbf{R@15\(^\circ\)}   & \textbf{R@30\(^\circ\)}   & \textbf{AUC\(_{30}\uparrow\)} \\ \midrule
     \textbf{DUSt3R} & \(--\)                     & 18.14 & 40.34 & 71.25 & 82.99 & 61.98 \\
     \textbf{Ours}   &DynamiCrafter          & \textbf{11.40}  & \textbf{55.21} & \textbf{89.93} &\textbf{93.75} & \textbf{77.41} \\
      \textbf{Ours}   &ToonCrafter            & 12.08 & 53.99 & 87.16 & 89.19 & 73.11 \\ 
      \bottomrule
      \end{tabular}
        }
      \end{table}

\begin{table}[!ht]
  \caption{Ablation study on the number of intermediate frames selected by FMS.}
  \label{tab:ab4}
  \centering
  \renewcommand{\arraystretch}{0.95}
  \setlength{\tabcolsep}{4pt}
  \scalebox{0.92}{
    \begin{tabular}{c c c c c c}
      \toprule
      \textbf{\#Frames} & \textbf{MRE\(\downarrow\)} & \textbf{R@5\(^\circ\)} & \textbf{R@15\(^\circ\)} & \textbf{R@30\(^\circ\)} & \textbf{AUC\(_{30}\uparrow\)} \\
      \midrule
      4 & 11.36 & 55.90 & \textbf{89.93} & 93.40 & 77.59 \\
      6 & \textbf{11.40} & 55.21 & \textbf{89.93} & \textbf{93.75} & \textbf{77.41} \\
      8 & 11.93 & \textbf{55.90} & \textbf{89.93} & 92.10 & 76.23 \\
      \bottomrule
    \end{tabular}
  }
\end{table}

\subsubsection{Qualitative Comparison on PoseCrafter}
\begin{figure}
    \centering
    \includegraphics[width=1\linewidth]{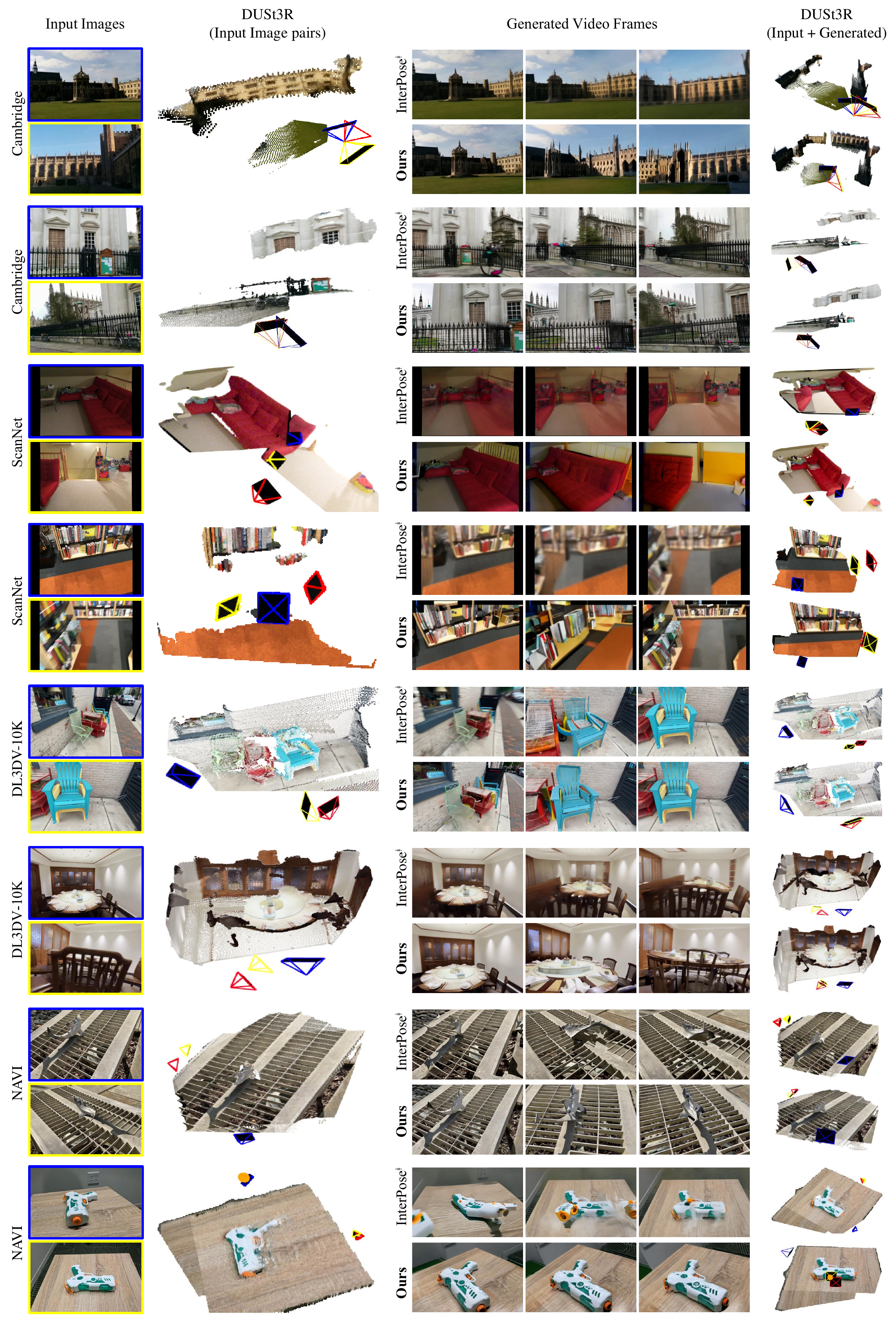}
    \caption{Qualitative comparisons on common benchmarks. \textcolor{blue}{Blue} and \textcolor{yellow}{yellow} outlines indicate the start and end frames, respectively. In the  DUSt3R reconstruction results, the visualized poses in \textcolor{blue}{blue}, \textcolor{yellow}{yellow}, and \textcolor{red}{red} represent the predicted pose of the start frame, predicted pose of the end frame, and the ground-truth pose of the end frame, respectively.}
    \label{fig2}
\end{figure}

To intuitively assess the effectiveness of our approach, we present qualitative comparisons with  DUSt3R and InterPose\(^{\ddagger}\) on the previously mentioned benchmarks in Figure~\ref{fig2}.
In Figure~\ref{fig2}, we visualize the input image pairs (with the start frame outlined in \textcolor{blue}{blue} and the end frame in \textcolor{yellow}{yellow}), point clouds reconstructed by  DUSt3R using only the input pairs, intermediate videos generated by InterPose\(^{\ddagger}\) and PoseCrafter, and the corresponding estimated poses and reconstructions produced by  DUSt3R for each video, respectively. 
Compared to  DUSt3R, which only uses input image pairs, PoseCrafter produces significantly cleaner and more complete geometry.
In addition, compared to InterPose\(^{\ddagger}\), our method is able to correct blur and pose errors in the middle of the sequence, reducing drift and artifacts in the reconstructed point cloud, thus providing more reliable visual evidence for subsequent pose estimation.
\subsubsection{Runtime and Memory Cost Discussion}
Following the quantitative and qualitative evaluations, we next analyze the efficiency of our approach. Specifically, we compare runtime and memory cost with InterPose\(^{\ddagger}\). 
InterPose generates 8 video sequences for each image pair and samples 11 frames from each sequence to compute the self-consistency score, resulting in a computationally heavy pipeline. 
In contrast, our method adopts a streamlined design: we generate only a single hybrid video and apply a deterministic frame selection strategy based on feature matching. 
This not only avoids redundant video synthesis but also eliminates repeated self-consistency evaluations. 

As summarized in Table~\ref{tab:cost}, our approach substantially reduces runtime in the pose estimation stage while maintaining comparable or even superior accuracy. 
Although the hybrid video generation incurs slightly higher memory usage than InterPose, the overall cost remains affordable for practical deployment. 
These results demonstrate that our framework strikes a favorable balance between efficiency and accuracy, underscoring its scalability to real-world applications.

\subsection{Ablation Studies}
In this section, we present ablation studies on the two key components of our framework: Hybrid Video Generation and the Feature Matching Selector. We further investigate the impact of different video generation models and the number of selected frames. All experiments are conducted on the Cambridge Landmarks benchmark under yaw changes of [50\(^{\circ}\)-65\(^{\circ}\)].
\paragraph{Hybrid Video Generation}
Table~\ref{tab:ab1} evaluates pose estimation performance for three video‐generation variants without any frame selection: DynamiCrafter (InterPose\(_{\mathrm{w/o\,SCS}}^{\ddagger}\)), ViewCrafter (Viewcrafter\(_{\mathrm{w/o\,FMS}}\)), and our hybrid video generation (Ours\(_{\mathrm{w/o\,FMS}}\)). Ours\(_{\mathrm{w/o\,FMS}}\) achieves the lower mean rotation error and higher recall than generated results from DynamiCrafter and ViewCrafter. 
This confirms that our hybrid video generation strategy, combining the motion-driven interpolation of DynamiCrafter with the pose-conditioned novel-view synthesis of ViewCrafter, effectively addresses the limitations of each method, enhancing the fidelity of intermediate frames for improved pose estimation.
\paragraph{Frame Matching Selector}
In Table~\ref{tab:ab2}, we evaluate the performance of different keypoint selection algorithms within the Frame Matching Selector (FMS), including RoMa~\citep{roma}, LoFTR~\citep{loftr}, ORB~\citep{orb}, SIFT~\citep{sift}, and SuperPoint~\citep{superpoint}.
We observe that ORB performs best in our case, which may be attributed to its robustness against color blurring and distortions in synthesized frames, making it more effective to select the most informative frames.
\paragraph{Impact of Video Interpolation Model}
To further validate the generality of our pipeline, we experimented with ToonCrafter~\citep{tooncrafter}, a recent interpolation model designed for cartoon-style data but also known for handling large-motion scenarios efficiently. 
As shown in Table~\ref{tab:ab3}, ToonCrafter performs slightly worse than DynamiCrafter, which is likely due to its training bias towards non-photorealistic content. 
Nevertheless, it still achieves clear improvements over directly using the input image pair. These results confirm that our hybrid pose estimation framework is compatible with a variety of video interpolation backbones, highlighting its flexibility and robustness.
\paragraph{Impact of Selected Frame Count}
Table~\ref{tab:ab4} analyzes the impact of varying the number of frames \(k\) selected by the FMS module, where \(k\) is 4, 6, and 8 (excluding the input image pair). The results show that \(k = 6\) yields the best overall performance. Using fewer frames may not provide sufficient information for accurate pose estimation, while using more frames can introduce errors due to artifacts in the generated frames.
\section{Limitation}
Although the effectiveness of our method has been validated across multiple diverse benchmarks, there are still certain cases where it may not perform well. For example, in our HVG stage, both the adopted video interpolation model (DynamiCrafter) and the novel view synthesis model (ViewCrafter) face challenges when the start and end frames exhibit significant illumination differences, resulting in synthesized frames with low confidence for subsequent pose estimation. This issue could potentially be mitigated by integrating a relighting model, such as IC-Light~\citep{zhang2025scaling}, to harmonize lighting conditions. In addition, our FMS module also encounters difficulties when the image texture is overly uniform, which hinders reliable feature matching. We will explore these limitations in future work.

\section{Conclusion}
In this work, we propose PoseCrafter, a simple yet effective framework for estimating camera poses from input image pairs with small or no overlap—commonly referred to as extreme viewpoint changes~\citep{InterPose}. Estimating poses under such challenging conditions remains a difficult problem, as minimal shared visual content often leads to poor matching and unreliable geometry. Rather than relying solely on video interpolation to synthesize intermediate frames, we introduce Hybrid Video Generation (HVG), which produces clearer and more geometrically consistent frames. HVG couples the vanilla interpolation model DynamiCrafter with a pose-conditioned novel view synthesis model, ViewCrafter, allowing it to refine and correct structural inconsistencies and reduce mid-sequence blur.
To further improve pose estimation accuracy, we also introduce Feature Matching Selector (FMS). This module identifies the most informative frames by evaluating their feature correspondences with the input image pair, ensuring that only geometrically meaningful frames suitable for pose estimation are selected. Through extensive experiments on multiple widely used benchmarks~\citep{InterPose}, including challenging scenes with extreme viewpoint differences, we demonstrate that PoseCrafter consistently outperforms existing state-of-the-art methods in both accuracy and robustness.
\paragraph{Acknowledgement.} 
This research / project is supported by the National Research Foundation (NRF) Singapore, under its NRF-Investigatorship Programme (Award ID. NRF-NRFI09-0008), and the China Scholarship Council under Grant Number 202306290143. It is also supported by the National Engineering Laboratory for Integrated Aero-Space-Ground-Ocean Big Data Application Technology and by the National Natural Science Foundation of China (NSFC) under Grant Nos. U19B2037, 61901384, and 61971356.

{
    \small
    \bibliographystyle{unsrt}
    \bibliography{neurips_2025}
}


\section*{NeurIPS Paper Checklist}

\begin{enumerate}

\item {\bf Claims}
    \item[] Question: Do the main claims made in the abstract and introduction accurately reflect the paper's contributions and scope?
    \item[] Answer: \answerYes{} 
    \item[] Justification: The main claims made in the abstract and introduction accurately reflect the paper's contributions and scope.
    \item[] Guidelines:
    \begin{itemize}
        \item The answer NA means that the abstract and introduction do not include the claims made in the paper.
        \item The abstract and/or introduction should clearly state the claims made, including the contributions made in the paper and important assumptions and limitations. A No or NA answer to this question will not be perceived well by the reviewers. 
        \item The claims made should match theoretical and experimental results, and reflect how much the results can be expected to generalize to other settings. 
        \item It is fine to include aspirational goals as motivation as long as it is clear that these goals are not attained by the paper. 
    \end{itemize}

\item {\bf Limitations}
    \item[] Question: Does the paper discuss the limitations of the work performed by the authors?
    \item[] Answer: \answerYes{} 
    \item[] Justification: The paper discusses the limitations of the work performed by the authors in the experiments section. 
    \item[] Guidelines:
    \begin{itemize}
        \item The answer NA means that the paper has no limitation while the answer No means that the paper has limitations, but those are not discussed in the paper. 
        \item The authors are encouraged to create a separate "Limitations" section in their paper.
        \item The paper should point out any strong assumptions and how robust the results are to violations of these assumptions (e.g., independence assumptions, noiseless settings, model well-specification, asymptotic approximations only holding locally). The authors should reflect on how these assumptions might be violated in practice and what the implications would be.
        \item The authors should reflect on the scope of the claims made, e.g., if the approach was only tested on a few datasets or with a few runs. In general, empirical results often depend on implicit assumptions, which should be articulated.
        \item The authors should reflect on the factors that influence the performance of the approach. For example, a facial recognition algorithm may perform poorly when image resolution is low or images are taken in low lighting. Or a speech-to-text system might not be used reliably to provide closed captions for online lectures because it fails to handle technical jargon.
        \item The authors should discuss the computational efficiency of the proposed algorithms and how they scale with dataset size.
        \item If applicable, the authors should discuss possible limitations of their approach to address problems of privacy and fairness.
        \item While the authors might fear that complete honesty about limitations might be used by reviewers as grounds for rejection, a worse outcome might be that reviewers discover limitations that aren't acknowledged in the paper. The authors should use their best judgment and recognize that individual actions in favor of transparency play an important role in developing norms that preserve the integrity of the community. Reviewers will be specifically instructed to not penalize honesty concerning limitations.
    \end{itemize}

\item {\bf Theory Assumptions and Proofs}
    \item[] Question: For each theoretical result, does the paper provide the full set of assumptions and a complete (and correct) proof?
    \item[] Answer: \answerNA{} 
    \item[] Justification: The theory assumptions in this paper have already been proved in previous works (Diffusion Model).
    \item[] Guidelines:
    \begin{itemize}
        \item The answer NA means that the paper does not include theoretical results. 
        \item All the theorems, formulas, and proofs in the paper should be numbered and cross-referenced.
        \item All assumptions should be clearly stated or referenced in the statement of any theorems.
        \item The proofs can either appear in the main paper or the supplemental material, but if they appear in the supplemental material, the authors are encouraged to provide a short proof sketch to provide intuition. 
        \item Inversely, any informal proof provided in the core of the paper should be complemented by formal proofs provided in appendix or supplemental material.
        \item Theorems and Lemmas that the proof relies upon should be properly referenced. 
    \end{itemize}

    \item {\bf Experimental Result Reproducibility}
    \item[] Question: Does the paper fully disclose all the information needed to reproduce the main experimental results of the paper to the extent that it affects the main claims and/or conclusions of the paper (regardless of whether the code and data are provided or not)?
    \item[] Answer: \answerYes{} 
    \item[] Justification: Details of training and evaluation are described in the Implementation Details section.
    \item[] Guidelines:
    \begin{itemize}
        \item The answer NA means that the paper does not include experiments.
        \item If the paper includes experiments, a No answer to this question will not be perceived well by the reviewers: Making the paper reproducible is important, regardless of whether the code and data are provided or not.
        \item If the contribution is a dataset and/or model, the authors should describe the steps taken to make their results reproducible or verifiable. 
        \item Depending on the contribution, reproducibility can be accomplished in various ways. For example, if the contribution is a novel architecture, describing the architecture fully might suffice, or if the contribution is a specific model and empirical evaluation, it may be necessary to either make it possible for others to replicate the model with the same dataset, or provide access to the model. In general. releasing code and data is often one good way to accomplish this, but reproducibility can also be provided via detailed instructions for how to replicate the results, access to a hosted model (e.g., in the case of a large language model), releasing of a model checkpoint, or other means that are appropriate to the research performed.
        \item While NeurIPS does not require releasing code, the conference does require all submissions to provide some reasonable avenue for reproducibility, which may depend on the nature of the contribution. For example
        \begin{enumerate}
            \item If the contribution is primarily a new algorithm, the paper should make it clear how to reproduce that algorithm.
            \item If the contribution is primarily a new model architecture, the paper should describe the architecture clearly and fully.
            \item If the contribution is a new model (e.g., a large language model), then there should either be a way to access this model for reproducing the results or a way to reproduce the model (e.g., with an open-source dataset or instructions for how to construct the dataset).
            \item We recognize that reproducibility may be tricky in some cases, in which case authors are welcome to describe the particular way they provide for reproducibility. In the case of closed-source models, it may be that access to the model is limited in some way (e.g., to registered users), but it should be possible for other researchers to have some path to reproducing or verifying the results.
        \end{enumerate}
    \end{itemize}

\item {\bf Open access to data and code}
    \item[] Question: Does the paper provide open access to the data and code, with sufficient instructions to faithfully reproduce the main experimental results, as described in supplemental material?
    \item[] Answer: \answerNo{} 
    \item[] Justification: We will open my code if our work is accepted. 
    \item[] Guidelines:
    \begin{itemize}
        \item The answer NA means that paper does not include experiments requiring code.
        \item Please see the NeurIPS code and data submission guidelines (\url{https://nips.cc/public/guides/CodeSubmissionPolicy}) for more details.
        \item While we encourage the release of code and data, we understand that this might not be possible, so “No” is an acceptable answer. Papers cannot be rejected simply for not including code, unless this is central to the contribution (e.g., for a new open-source benchmark).
        \item The instructions should contain the exact command and environment needed to run to reproduce the results. See the NeurIPS code and data submission guidelines (\url{https://nips.cc/public/guides/CodeSubmissionPolicy}) for more details.
        \item The authors should provide instructions on data access and preparation, including how to access the raw data, preprocessed data, intermediate data, and generated data, etc.
        \item The authors should provide scripts to reproduce all experimental results for the new proposed method and baselines. If only a subset of experiments are reproducible, they should state which ones are omitted from the script and why.
        \item At submission time, to preserve anonymity, the authors should release anonymized versions (if applicable).
        \item Providing as much information as possible in supplemental material (appended to the paper) is recommended, but including URLs to data and code is permitted.
    \end{itemize}

\item {\bf Experimental Setting/Details}
    \item[] Question: Does the paper specify all the training and test details (e.g., data splits, hyperparameters, how they were chosen, type of optimizer, etc.) necessary to understand the results?
    \item[] Answer: \answerYes{} 
    \item[] Justification: The paper specify all the training and test details.
    \item[] Guidelines:
    \begin{itemize}
        \item The answer NA means that the paper does not include experiments.
        \item The experimental setting should be presented in the core of the paper to a level of detail that is necessary to appreciate the results and make sense of them.
        \item The full details can be provided either with the code, in appendix, or as supplemental material.
    \end{itemize}

\item {\bf Experiment Statistical Significance}
    \item[] Question: Does the paper report error bars suitably and correctly defined or other appropriate information about the statistical significance of the experiments?
    \item[] Answer: \answerYes{}{} 
    \item[] Justification: The paper adopts the mean metric values for 3 times testing for a fair comparison. 
    \item[] Guidelines:
    \begin{itemize}
        \item The answer NA means that the paper does not include experiments.
        \item The authors should answer "Yes" if the results are accompanied by error bars, confidence intervals, or statistical significance tests, at least for the experiments that support the main claims of the paper.
        \item The factors of variability that the error bars are capturing should be clearly stated (for example, train/test split, initialization, random drawing of some parameter, or overall run with given experimental conditions).
        \item The method for calculating the error bars should be explained (closed form formula, call to a library function, bootstrap, etc.)
        \item The assumptions made should be given (e.g., Normally distributed errors).
        \item It should be clear whether the error bar is the standard deviation or the standard error of the mean.
        \item It is OK to report 1-sigma error bars, but one should state it. The authors should preferably report a 2-sigma error bar than state that they have a 96\% CI, if the hypothesis of Normality of errors is not verified.
        \item For asymmetric distributions, the authors should be careful not to show in tables or figures symmetric error bars that would yield results that are out of range (e.g. negative error rates).
        \item If error bars are reported in tables or plots, The authors should explain in the text how they were calculated and reference the corresponding figures or tables in the text.
    \end{itemize}

\item {\bf Experiments Compute Resources}
    \item[] Question: For each experiment, does the paper provide sufficient information on the computer resources (type of compute workers, memory, time of execution) needed to reproduce the experiments?
    \item[] Answer: \answerYes{} 
    \item[] Justification: In the Experiments section Implementation  Details section, the paper introduces the detailed implementation details, including the computing resources, training and evaluation details. 
    \item[] Guidelines:
    \begin{itemize}
        \item The answer NA means that the paper does not include experiments.
        \item The paper should indicate the type of compute workers CPU or GPU, internal cluster, or cloud provider, including relevant memory and storage.
        \item The paper should provide the amount of compute required for each of the individual experimental runs as well as estimate the total compute. 
        \item The paper should disclose whether the full research project required more compute than the experiments reported in the paper (e.g., preliminary or failed experiments that didn't make it into the paper). 
    \end{itemize}
    
\item {\bf Code Of Ethics}
    \item[] Question: Does the research conducted in the paper conform, in every respect, with the NeurIPS Code of Ethics \url{https://neurips.cc/public/EthicsGuidelines}?
    \item[] Answer: \answerYes{} 
    \item[] Justification: The paper is sure to preserve anonymity.
    \item[] Guidelines: 
    \begin{itemize}
        \item The answer NA means that the authors have not reviewed the NeurIPS Code of Ethics.
        \item If the authors answer No, they should explain the special circumstances that require a deviation from the Code of Ethics.
        \item The authors should make sure to preserve anonymity (e.g., if there is a special consideration due to laws or regulations in their jurisdiction).
    \end{itemize}

\item {\bf Broader Impacts}
    \item[] Question: Does the paper discuss both potential positive societal impacts and negative societal impacts of the work performed?
    \item[] Answer: \answerYes{} 
    \item[] Justification: This paper contributes to extreme pose estimation research and extends beyond just extreme pose estimation tasks. In the future, it has the potential to enhance other research fields and benefit human society.
    \item[] Guidelines:
    \begin{itemize}
        \item The answer NA means that there is no societal impact of the work performed.
        \item If the authors answer NA or No, they should explain why their work has no societal impact or why the paper does not address societal impact.
        \item Examples of negative societal impacts include potential malicious or unintended uses (e.g., disinformation, generating fake profiles, surveillance), fairness considerations (e.g., deployment of technologies that could make decisions that unfairly impact specific groups), privacy considerations, and security considerations.
        \item The conference expects that many papers will be foundational research and not tied to particular applications, let alone deployments. However, if there is a direct path to any negative applications, the authors should point it out. For example, it is legitimate to point out that an improvement in the quality of generative models could be used to generate deepfakes for disinformation. On the other hand, it is not needed to point out that a generic algorithm for optimizing neural networks could enable people to train models that generate Deepfakes faster.
        \item The authors should consider possible harms that could arise when the technology is being used as intended and functioning correctly, harms that could arise when the technology is being used as intended but gives incorrect results, and harms following from (intentional or unintentional) misuse of the technology.
        \item If there are negative societal impacts, the authors could also discuss possible mitigation strategies (e.g., gated release of models, providing defenses in addition to attacks, mechanisms for monitoring misuse, mechanisms to monitor how a system learns from feedback over time, improving the efficiency and accessibility of ML).
    \end{itemize}
    
\item {\bf Safeguards}
    \item[] Question: Does the paper describe safeguards that have been put in place for responsible release of data or models that have a high risk for misuse (e.g., pretrained language models, image generators, or scraped datasets)?
    \item[] Answer: \answerNA{} 
    \item[] Justification: The paper poses no such risks.
    \item[] Guidelines:
    \begin{itemize}
        \item The answer NA means that the paper poses no such risks.
        \item Released models that have a high risk for misuse or dual-use should be released with necessary safeguards to allow for controlled use of the model, for example by requiring that users adhere to usage guidelines or restrictions to access the model or implementing safety filters. 
        \item Datasets that have been scraped from the Internet could pose safety risks. The authors should describe how they avoided releasing unsafe images.
        \item We recognize that providing effective safeguards is challenging, and many papers do not require this, but we encourage authors to take this into account and make a best faith effort.
    \end{itemize}

\item {\bf Licenses for existing assets}
    \item[] Question: Are the creators or original owners of assets (e.g., code, data, models), used in the paper, properly credited and are the license and terms of use explicitly mentioned and properly respected?
    \item[] Answer: \answerYes{} 
    \item[] Justification: The paper cites the original paper that produced the code package or dataset, and follows the licenses for existing assets.
    \item[] Guidelines:
    \begin{itemize}
        \item The answer NA means that the paper does not use existing assets.
        \item The authors should cite the original paper that produced the code package or dataset.
        \item The authors should state which version of the asset is used and, if possible, include a URL.
        \item The name of the license (e.g., CC-BY 4.0) should be included for each asset.
        \item For scraped data from a particular source (e.g., website), the copyright and terms of service of that source should be provided.
        \item If assets are released, the license, copyright information, and terms of use in the package should be provided. For popular datasets, \url{paperswithcode.com/datasets} has curated licenses for some datasets. Their licensing guide can help determine the license of a dataset.
        \item For existing datasets that are re-packaged, both the original license and the license of the derived asset (if it has changed) should be provided.
        \item If this information is not available online, the authors are encouraged to reach out to the asset's creators.
    \end{itemize}

\item {\bf New Assets}
    \item[] Question: Are new assets introduced in the paper well documented and is the documentation provided alongside the assets?
    \item[] Answer: \answerNA{} 
    \item[] Justification: This paper does not introduce new assets.
    \item[] Guidelines:
    \begin{itemize}
        \item The answer NA means that the paper does not release new assets.
        \item Researchers should communicate the details of the dataset/code/model as part of their submissions via structured templates. This includes details about training, license, limitations, etc. 
        \item The paper should discuss whether and how consent was obtained from people whose asset is used.
        \item At submission time, remember to anonymize your assets (if applicable). You can either create an anonymized URL or include an anonymized zip file.
    \end{itemize}

\item {\bf Crowdsourcing and Research with Human Subjects}
    \item[] Question: For crowdsourcing experiments and research with human subjects, does the paper include the full text of instructions given to participants and screenshots, if applicable, as well as details about compensation (if any)? 
    \item[] Answer: \answerNA{} 
    \item[] Justification: The paper does not involve crowdsourcing nor research with human subjects.
    \item[] Guidelines:
    \begin{itemize}
        \item The answer NA means that the paper does not involve crowdsourcing nor research with human subjects.
        \item Including this information in the supplemental material is fine, but if the main contribution of the paper involves human subjects, then as much detail as possible should be included in the main paper. 
        \item According to the NeurIPS Code of Ethics, workers involved in data collection, curation, or other labor should be paid at least the minimum wage in the country of the data collector. 
    \end{itemize}

\item {\bf Institutional Review Board (IRB) Approvals or Equivalent for Research with Human Subjects}
    \item[] Question: Does the paper describe potential risks incurred by study participants, whether such risks were disclosed to the subjects, and whether Institutional Review Board (IRB) approvals (or an equivalent approval/review based on the requirements of your country or institution) were obtained?
    \item[] Answer: \answerNA{} 
    \item[] Justification: The paper does not involve crowdsourcing nor research with human subjects.
    \item[] Guidelines: 
    \begin{itemize}
        \item The answer NA means that the paper does not involve crowdsourcing nor research with human subjects.
        \item Depending on the country in which research is conducted, IRB approval (or equivalent) may be required for any human subjects research. If you obtained IRB approval, you should clearly state this in the paper. 
        \item We recognize that the procedures for this may vary significantly between institutions and locations, and we expect authors to adhere to the NeurIPS Code of Ethics and the guidelines for their institution. 
        \item For initial submissions, do not include any information that would break anonymity (if applicable), such as the institution conducting the review.
    \end{itemize}

\end{enumerate}


\newpage
\appendix

\section{Camera Pose Estimation Results based on MASt3R}
To further validate the generality of PoseCrafter on different pose estimation backbones, we replaced DUSt3R~\citep{DUSt3R} with MASt3R~\citep{mast3r}, another transformer‐based framework augmented with a dense local feature head and efficient sparse matching. 
We evaluated this configuration on four benchmarks with varying yaw ranges: Cambridge Landmarks \((50^\circ\text{--}65^\circ)\), ScanNet \((50^\circ\text{--}65^\circ)\), DL3DV-10K \((50^\circ\text{--}90^\circ)\), and NAVI \((50^\circ\text{--}90^\circ)\).
As shown in Table~\ref{tab:cam&scannet} and Table~\ref{tab:dl3dv&navi}, using MASt3R directly for pose estimation results in lower accuracy compared to DUSt3R. A possible explanation is that the sparse matching algorithm employed by MASt3R may be less effective for image pairs with small or no overlap. Nonetheless, our proposed framework consistently improves estimation accuracy over the baseline pose estimation models, including MASt3R. This demonstrates that our approach generalizes well across different pose estimation backbones by synthesizing intermediate frames that are more suitable for pose estimation.

\begin{table}[!ht]
  \caption{Camera pose estimation results on outward-facing datasets (Cambridge Landmarks and ScanNet) based on MASt3R. We report rotation recall (R@\(\,\theta\uparrow\)), translation recall (T@\(\,\theta\uparrow\)), mean rotation error (MRE\(\downarrow\)), mean translation error (MTE\(\downarrow\)), and AUC\(_{30}\uparrow\).}
  \label{tab:cam&scannet}
  \centering
\resizebox{\linewidth}{!}{
\begin{tabular}{llccccccccccccccc}
  \toprule
    & & \multicolumn{5}{c}{ \textbf{Cambridge Landmarks}}  &   & \multicolumn{9}{c}{\textbf{ScanNet}}  \\ \cmidrule(lr){3-7}  \cmidrule(lr){9-17}
\multirow{-2}{*}{\textbf{Method}} & \multirow{-2}{*}{\textbf{Input}} &  \textbf{R@5\(^\circ\)} & \textbf{R@15\(^\circ\)} & \textbf{R@30\(^\circ\)} &  \textbf{MRE}   &  \textbf{AUC\(_{30}\)} &         &  \textbf{R@5\(^\circ\)} & \textbf{R@15\(^\circ\)} & \textbf{R@30\(^\circ\)} & \textbf{T@5\(^\circ\)} & \textbf{T@15\(^\circ\)} & \textbf{T@30\(^\circ\)} &  \textbf{MRE}   &  \textbf{MTE}   &  \textbf{AUC\(_{30}\)} \\ \midrule
 MASt3R &  Pair   & 9.03 & 43.75 & 57.99  & 51.61 & 36.15 &   &25.00   &55.17 &62.93 &6.90 &30.17  &48.28  &36.34 &50.13 &26.84 \\
 Ours(MASt3R) &  Hybrid video & \textbf{25.00} & \textbf{59.03} &\textbf{69.10}  &\textbf{41.02} & \textbf{49.83} &  &\textbf{31.90}   &\textbf{65.52} &\textbf{77.59} &\textbf{18.97}  &\textbf{45.69} &\textbf{63.79}   &\textbf{29.97} &\textbf{38.50} &\textbf{40.66} \\  \bottomrule
\end{tabular}}
\end{table}

\begin{table}[!ht]
  \caption{Camera pose estimation results on center-facing datasets (DL3DV-10K and NAVI) based on MASt3R. We report rotation recall (R@\(\,\theta\uparrow\)), translation recall (T@\(\,\theta\uparrow\)), mean rotation error (MRE\(\downarrow\)), mean translation error (MTE\(\downarrow\)), and AUC\(_{30}\uparrow\).}
  \vspace{0.05in}
  \label{tab:dl3dv&navi}
  \centering
  \resizebox{\linewidth}{!}{
    \begin{tabular}{l l l c c c c c c c c c c}
      \toprule
      \textbf{Dataset} & \textbf{Method} & \textbf{Input} & \textbf{R@5\(^\circ\)} & \textbf{R@15\(^\circ\)} & \textbf{R@30\(^\circ\)} & \textbf{T@5\(^\circ\)} & \textbf{T@15\(^\circ\)} & \textbf{T@30\(^\circ\)} & \textbf{MRE} & \textbf{MTE} & \textbf{AUC\(_{30}\)} \\
      \midrule
      \multirow{2}{*}{DL3DV-10K}& MASt3R                         & Pair            & 7.00                              & 63.67                              & 97.00                              & 26.67                             & 72.67                              & 91.67                              & 15.18          & 13.31          & 53.49           \\
      &Ours(MASt3R)   & Hybrid video & \textbf{7.67}                     & \textbf{68.67}                     & \textbf{97.33}                     & \textbf{30.00}                    & \textbf{76.33}                     & \textbf{92.00}                     & \textbf{14.09} & \textbf{12.91} & \textbf{55.36}  \\
      \midrule
      \multirow{2}{*}{NAVI}&MASt3R            & Pair  & \textbf{43.97}                    & 93.39                              & 96.89                              & 50.97                             & 89.49                              & \textbf{97.67}                     & 8.30            & 7.43           & 76.54 \\
      &Ours(MASt3R)        & Hybrid video    & \textbf{43.97}                    & \textbf{93.77}                     & \textbf{97.28}                     & \textbf{51.75}                    & \textbf{91.83}                     & 97.28                              & \textbf{7.71}  & \textbf{6.98}  & \textbf{77.26} \\
      \bottomrule
    \end{tabular}
  }
\end{table}

\begin{table}[!ht]
  \caption{Additional comparison with VGGT on Cambridge Landmarks. We report mean rotation error (MRE\(\downarrow\)), rotation recall (R@\(\,\theta\uparrow\)), and AUC\(_{30}\uparrow\).}
  \vspace{0.05in}
  \label{tab:vggt}
  \resizebox{\linewidth}{!}{
\begin{tabular}{lcccccc}
\toprule
\textbf{Method}     & \textbf{Input}        & \textbf{MRE}   & \textbf{R@5\(^\circ\)}    & \textbf{R@15\(^\circ\)}   & \textbf{R@30\(^\circ\)}   & \textbf{AUC\(_{30}\)}           \\ \midrule
Dust3R     & Pair         & 18.14          & 40.34          & 71.25          & 82.99          & 61.98          \\
Ours       & Hybrid video & \textbf{11.40} & \textbf{55.21} & \textbf{89.93} & \textbf{93.75} & \textbf{77.41} \\
VGGT       & Pair         & 20.17          & 40.00          & 70.17          & 82.29          & 60.54          \\
VGGT\(_{Ours}\) & Hybrid video & 17.88          & 42.43          & 84.40          & 85.76          & 65.15          \\ \bottomrule
\end{tabular}}
\end{table}
\section{Additional Comparison with VGGT}
To further analyze the generality of our pipeline, we conducted comparative experiments with VGGT~\cite{vggt} on the Cambridge Landmarks dataset under yaw changes of [50\(^{\circ}\)–65\(^{\circ}\)]. 

We first evaluated the two models in challenging cases where only image pairs with small or no overlap were provided. As shown in the second and fourth rows of Table~\ref{tab:vggt}, DUSt3R consistently outperforms VGGT on such data. 

We then integrate these two models into our pipeline and evaluate their performance. As shown in the third and fifth rows, both configurations achieve significant improvements over their directly estimated counterparts. The version using DUSt3R achieves a higher accuracy than the version using VGGT. These results demonstrate that our method is compatible with different pose estimators and consistently enhances their performance.

Overall, DUSt3R appears better suited than VGGT for small- or non-overlapping image pairs, and our framework provides more noticeable improvements when combined with DUSt3R. While fine-tuning VGGT on small-overlap data or adjusting the hyperparameters of our pipeline may further enhance its performance, we leave such extensions for future work.
\section{Comparison of DUSt3R Confidence-based Selection and FMS}
We compare DUSt3R confidence-based frame selection with our proposed Feature Matching Selector (FMS) to evaluate whether confidence scores can serve as a viable alternative. 
In this setting, top-ranked frames were selected from hybrid videos using four different confidence thresholds (20\%, 40\%, 60\%, and 80\%). The results are summarized in Table~\ref{tab:confidence}. 

When the threshold was set to 20\% or 40\%, pose estimation accuracy decreased compared to using the full hybrid video sequence. Increasing the threshold to 60\% or 80\% improved performance, but the accuracy still remained lower than that achieved by our proposed FMS. 

In terms of efficiency, confidence-based selection introduces substantial overhead, as all video frames must be processed by DUSt3R to compute confidence maps prior to selection. This step incurs significant time and memory costs, and higher thresholds further increase runtime as more frames are selected. 
By contrast, our FMS requires only a single feature extraction step, immediately identifying the most informative frames with superior efficiency and accuracy. These findings highlight the practical advantages of our proposed FMS over confidence-based selection in both computational efficiency and pose estimation performance.
\begin{table}[!ht]
\caption{Comparison of DUSt3R confidence-based selection and FMS on Cambridge Landmarks.
We report pose estimation results under yaw changes of [50\(^{\circ}\)–65\(^{\circ}\)]. 
Frames were selected from hybrid videos using four DUSt3R confidence thresholds (20\%, 40\%, 60\%, and 80\%).We report mean rotation error (MRE\(\downarrow\)), rotation recall (R@\(\,\theta\uparrow\)),AUC\(_{30}\uparrow\) and pose estimation time.}
  \vspace{0.05in}
  \label{tab:confidence}
  \resizebox{\linewidth}{!}{
\begin{tabular}{lcccccc}
\toprule
\textbf{Method}    &\textbf{MRE↓}   & \textbf{R@5\(^\circ\)}    & \textbf{R@15\(^\circ\)}   & \textbf{R@30\(^\circ\)}   & \textbf{AUC\(_{30}\uparrow\)} & \textbf{Pose Estimation Time} \\
\midrule
\textbf{Conf(20\(\%\))}         & 14.66 & 54.17 & 85.07 & 89.24 & 72.45 & 2.79min        \\ 
\textbf{Conf(40\(\%\))}         & 14.36 & 57.30 & 87.15 & 90.97 & 74.24 & 2.91min        \\ 
\textbf{Conf(60\(\%\))}        & 12.35 & 54.17 & 90.28 & 93.06 & 76.60 & 3.47min        \\ 
\textbf{Conf(80\(\%\))}         & 12.41 & 53.47 & 90.63 & 93.06 & 76.46 & 4.13min        \\ 
\textbf{Ours\(_{\text{w/o FMS}}\)}  & 13.24 & 54.51 & 89.24 & 92.71 & 76.13 & 2.56min        \\ 
\textbf{Ours}       & \textbf{11.40} & \textbf{55.21} & \textbf{89.93} & \textbf{93.75} & \textbf{77.41} & \textbf{0.18min}       \\ \bottomrule
\end{tabular}}
\end{table}
\section{Visualization Results of Limitation}
\begin{figure}[!ht]
            \centering
            \includegraphics[width=1\linewidth]{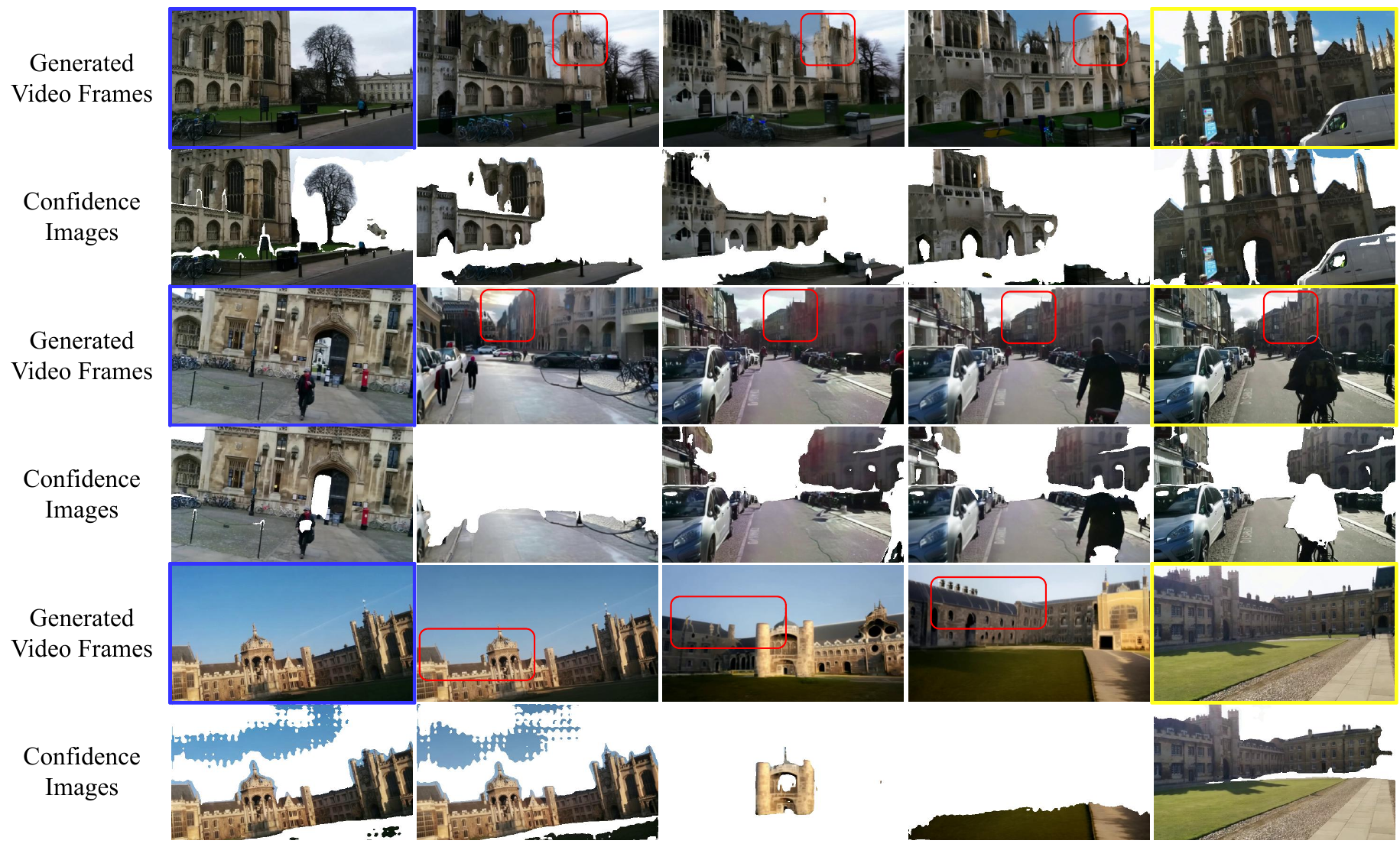}
            \caption{Hybrid Video Generation artifacts under severe illumination differences. \textcolor{blue}{Blue} and \textcolor{yellow}{yellow} outlines indicate the start and end frames, respectively. Confidence images denote images filtered with the predicted confidence map in the subsequent DUSt3R model.
            The \textcolor{red}{red} boxes highlight regions affected by artifacts in these cases. We can observe that these regions have quite low confidence.}
            \label{fig:limitation1}
\end{figure}
\begin{figure}[!ht]
            \centering
            \includegraphics[width=1\linewidth]{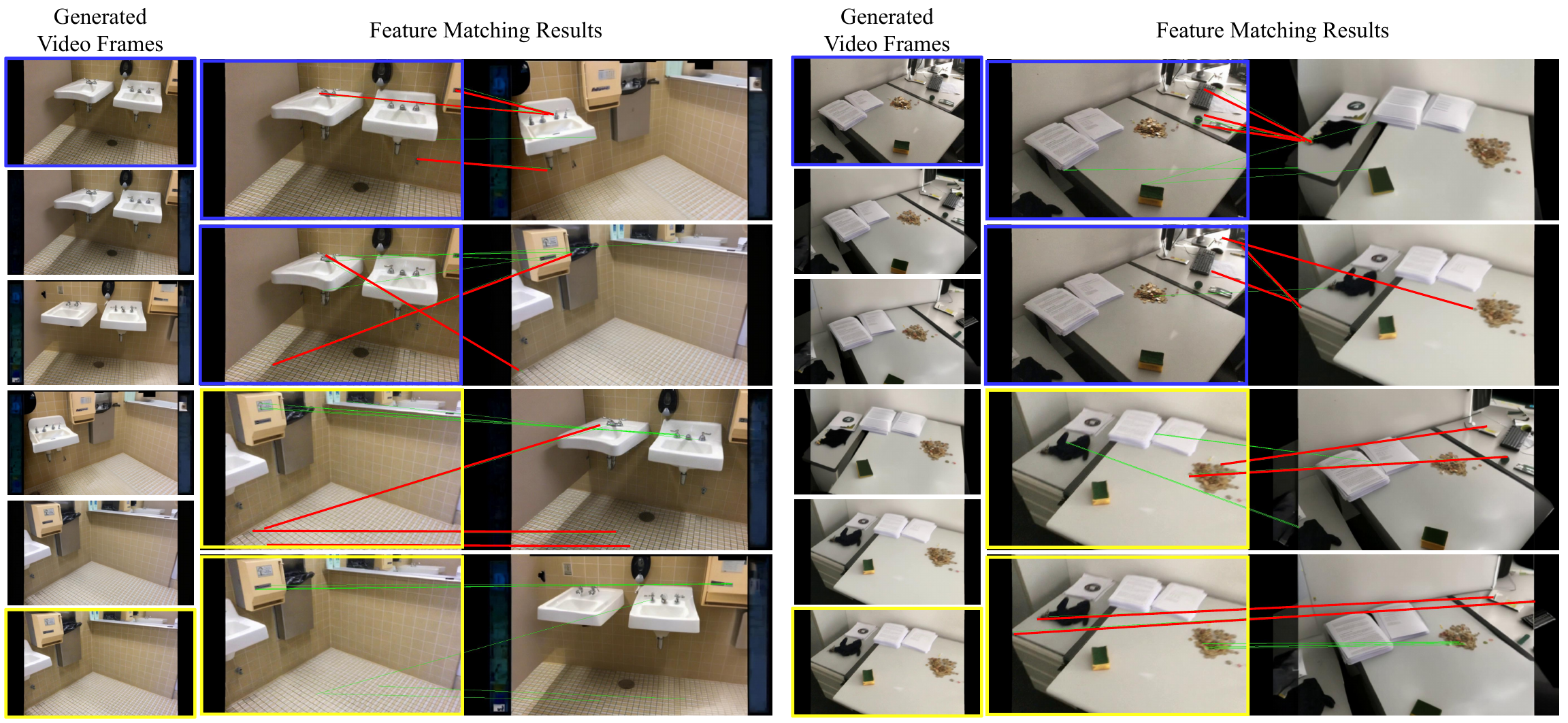}
            \caption{Feature Matching Selection failures in low-texture regions. \textcolor{blue}{Blue} and \textcolor{yellow}{yellow} outlines indicate the start and end frames, respectively. 
            The \textcolor{red} {red} and \textcolor{green}{green} lines indicate incorrect and correct correspondences, respectively. Incorrect correspondences lead to errors in inlier counting, affecting the accuracy of subsequent frame selection.
            }
            \label{fig:limitation2}
\end{figure}
Although PoseCrafter achieves robust results across various benchmarks, certain challenging scenes still degrade its performance. 
In the hybrid video generation (HVG) stage, severe illumination differences between the start and end frames will introduce obvious artifacts in the synthesized intermediate views.
The \textcolor{red}{red-boxed} regions in Figure~\ref{fig:limitation1} highlight these artifacts, which would be marked as low confidence and discarded in the downstream pose estimation backbone, e.g., DUSt3R in this work. Such an obvious removal of information may significantly degrade the final performance.
Moreover, in the feature match selector (FMS) module, scenes dominated by uniform or repetitive textures hinder the reliable extraction and matching of key points, resulting in fewer RANSAC~\citep{ransac} inliers and lower pose accuracy. The \textcolor{red}{red} lines in Figure \ref{fig:limitation2} show examples of incorrect correspondences in such scenes.

\section{More Visualizations of Our Generated Videos}
\begin{figure}
            \centering
            \includegraphics[width=1\linewidth]{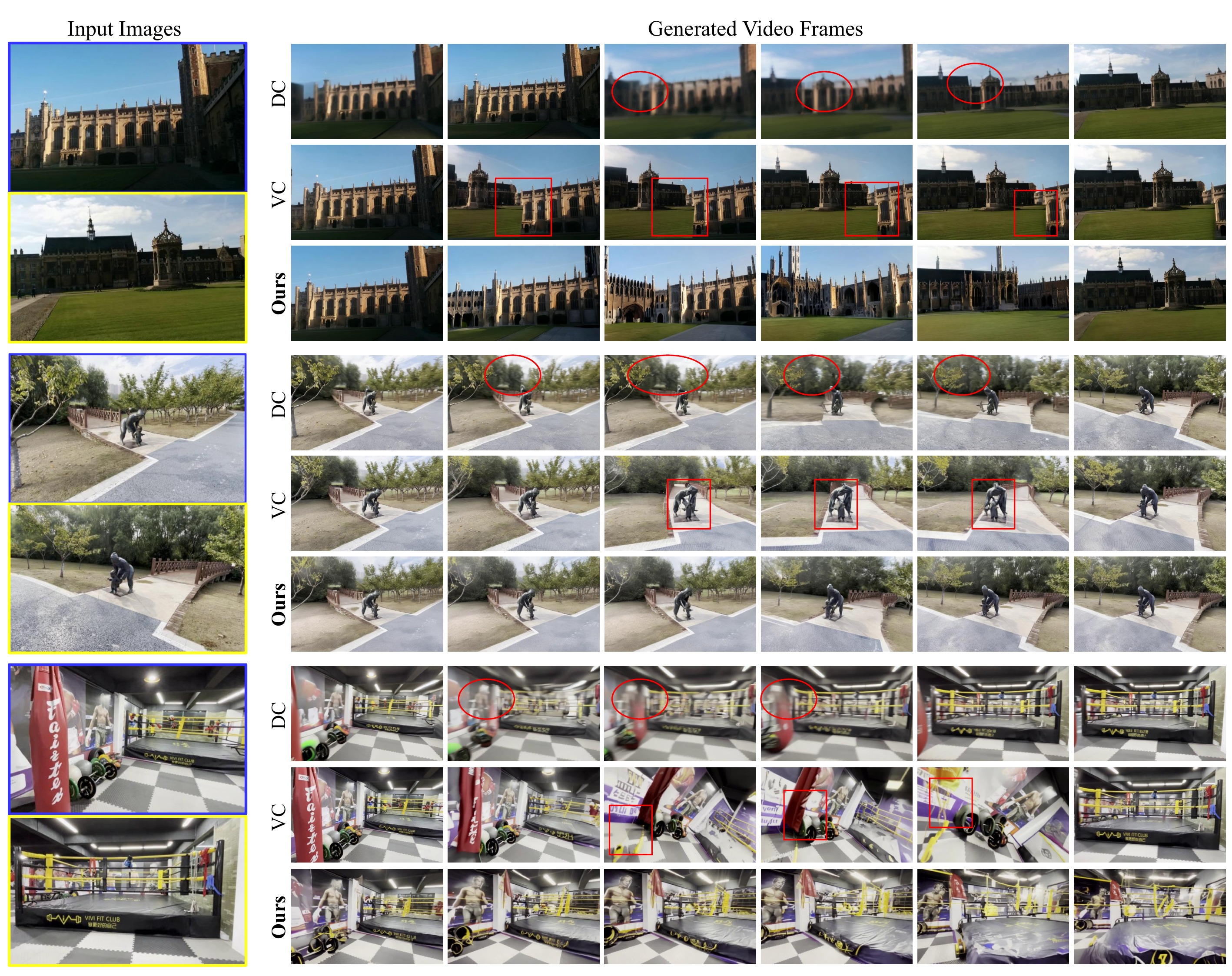}
            \caption{Comparative video synthesis results. Each row shows intermediate frames synthesized between the same start frame (\textcolor{blue}{blue} box) and end frame (\textcolor{yellow}{yellow} box), generated by different methods: DynamiCrafter (DC), ViewCrafter (VC), and our Hybrid Video Generation (Ours). DC produces smooth motion but exhibits progressive blur and geometric drift in the middle of sequences (highlighted in \textcolor{red} {red} circle).
            Since VC is sensitive to the pose of the input image pair, it tends to produce structural misalignments in our small-overlap setting (highlighted in \textcolor{red} {red} box).
            By coupling DC and VC together, our method delivers sharp, geometrically consistent video frames throughout, correcting both the blur of DC and the misalignments of VC.}
            \label{fig:video_compare}
            \vspace{-0.2in}
        \end{figure}

To further illustrate the superiority of our proposed hybrid video generation(HVG), we present additional side‐by‐side comparisons of intermediate frames produced by DynamiCrafter, ViewCrafter, and PoseCrafter. In Figure \ref{fig:video_compare}, DynamiCrafter~\citep{dynamicrafter} delivers temporally smooth transitions but exhibits progressive blur and geometric drift in the middle of generated sequences.
Although ViewCrafter~\citep{viewcrafter} can produce sharp results with minimal blur, using only input image pairs with small overlap often leads to structural artifacts and misalignments.
To address these limitations, we combine DynamiCrafter and ViewCrafter to complement each other's strengths. Specifically, we first use DynamiCrafter to synthesize intermediate “relay” frames, effectively augmenting the input image pair with frames that have larger overlaps. These relay frames are then passed to ViewCrafter to generate clearer and more geometrically consistent results. As shown in Figure~\ref{fig:video_compare}, our proposed approach successfully produces frames that are both visually sharp and structurally reliable.  

\end{document}